\pdfoutput=1

\documentclass[11pt]{article}

\usepackage{acl}
\usepackage{times}
\usepackage{latexsym}
\usepackage[T1]{fontenc}

\usepackage[utf8]{inputenc}
\usepackage{amsmath}
\usepackage{amssymb}
\usepackage{amsthm}
\usepackage{bbm}
\usepackage{graphicx}
\usepackage{arydshln}
\usepackage{relsize}
\usepackage{empheq}

\title{The Trade-offs of Domain Adaptation \\for Neural Language Models}

\author{%
David Grangier\\
Google, Mountain View, CA\\
\texttt{grangier@google.com}\\\And
Dan Iter\thanks{~~Work performed while interning at Google.}\\
Stanford, Palo Alto, CA\\
\texttt{daniter@stanford.edu}}

\DeclareMathOperator*{\E}{\mathbb{E}}

\DeclareMathOperator*{\argmin}{arg\,min}

\newcommand*\widefbox[1]{\fbox{\hspace{0em}#1\hspace{0em}}}

\begin{document}
\maketitle

\begin{abstract}
This work connects language model adaptation with concepts of machine learning theory.
We consider a training setup with a large out-of-domain set and a small in-domain set. We 
derive how the benefit of training a model on either set depends 
on the size of the sets and the distance between their underlying distributions. 
We analyze how out-of-domain pre-training before in-domain fine-tuning 
achieves better generalization than either solution independently. Finally, we 
present how adaptation techniques based on data selection, such as importance 
sampling, intelligent data selection and influence functions, can be presented in a 
common framework which highlights their similarity and also their subtle differences.
\end{abstract}

\section{Introduction}
Neural Language Models (LMs) trained on large generic training sets – over a billion 
sentences~\cite{kaplan2020scaling, roziewski2021languagecrawl} – have been shown to be effective 
at adapting to smaller, specific target domains for language modeling and other downstream tasks~\cite{foundation21}.
Neural LM adaptation is commonly performed via fine tuning ~\cite{bert,liu2019roberta,t5,gpt}, data 
selection ~\cite{van-der-wees-etal-2017-dynamic} or their combination ~\cite{wang-etal-2018-denoising,aharoni2020unsupervised,Gururangan2020}. 
However, the tradeoffs between fine-tuning and reweighting of pre-training data is not well understood and a theoretical framework for reasoning about the generalization performance of these methods is needed.

In this paper, we connect language model adaptation with concepts of machine learning theory. Our derivations support past empirical observations: it has been observed that the size of the out-of-domain pre-training set is important for in domain generalization~\cite{t5,bert} or that domain  adaptation is more effective on
domains which are well represented in the the pre-training data~\cite{gpt}.
Our study consider a training setup with a large out-of-domain set and a small in-domain set. As a
first contribution, we derive how the benefit of training a model on either set depends 
on the size of the sets and the distance between their underlying distribution. 
We also expose how fine-tuning can be viewed as a regularization method that can
achieve a better trade-off than training only on either set.

The research on data selection for LM adaption originates mainly from intelligent 
selection~\cite{moore-lewis-2010-intelligent,axelrod-etal-2011-domain}. This 
method examines the out-of-domain training data to emphasize a subset
deemed more likely by an in-domain model than by an out-of-domain model. 
Although intuitive, the connection of this method with statistical 
estimation is unclear, which makes studying its impact on generalization 
error difficult. Another family of selection methods stems from influence functions~\cite{koh-liang-influence-17,wang2021gradientguided}
which estimate whether the model updates from out-of-domain training examples are
aligned with the in-domain updates. This approach is more principled and its impact on 
the generalization error is easier to study. In this work, as a second contribution, we 
show how intelligent selection and influence function methods are linked in the 
case of neural LMs. In particular, we show that they both can be derived from 
importance sampling~\cite{owen13}, a classical, well-studied statistical estimation technique.

The rest of our paper is organized as follows. We first presents the theoretical trade-offs between in-domain and out-of-domain training. We highlight the importance of the relative sizes of in-domain and out-of-domain training sets along with the distance between their underlying distributions. We also present how fine-tuning with a limited number of updates can be seen as a training method regularized with respect to the out-of-domain prior. Finally, we present data selection methods under a unifying framework.

\section{Neural Language Modeling}

Language modeling refers to the generative modeling of natural language~\cite{manning1999foundations}. 
Commonly, natural language is represented as a sequence of symbols, tokens, 
from a finite vocabulary. For instance, language can be represented as a 
sequence of characters, a sequence of words or alternative units. A neural 
language model (LM) decomposes the estimates the log probability of a text $y = 
(y_1, \ldots y_n)$, as
$$
\log P(y; \theta) = \sum^{n}_{i=1} \log P(y_i|y^{i-1}_1; \theta)
$$
where $P_\theta$ maps a parameter vector $\theta$ along 
with a sequence of past tokens $y^{i-1}_1$ onto a probability distribution 
over the vocabulary. Different types of neural architectures have been used 
for neural language modeling. Most architectures used for LMs 
re-use intermediate computations from the previous steps for the next steps
when estimating probabilities for successive tokens in the same sequence.
Popular architectures include recurrent neural networks~\cite{DBLP:conf/interspeech/MikolovKBCK10,DBLP:conf/interspeech/SundermeyerSN12}, convolutional 
networks~\cite{DBLP:conf/icml/DauphinFAG17} and transformer networks~\cite{transformer,gpt}.

The parameter vector $\theta \in \Theta$ of a neural LM is 
identified by maximizing the log likelihood over a training set $D$ sampled
from the true distribution ${\cal D}$ using variants of stochastic gradient 
descent. The log likelihood of a held-out set, sampled from the same distribution, 
can evaluate model quality. One often reports perplexity,
the exponentiated negative average 
log likelihood per token.

Conditional LMs model the distribution of 
a text $y$ given a conditioning input $x$. 
$$
\log P(y|x; \theta) = \sum^{n}_{i=1} \log P(y_i|y^{i-1}_1, x; \theta)
$$
This type of model is used for translation where $(x, y)$ pairs are 
sentences in the source and target language~\cite{koehn2009statistical,DBLP:journals/corr/BahdanauCB14} 
or summarization where $(x, y)$ pairs are corresponding articles and summaries~\cite{DBLP:journals/corr/SeeLM17}.

For both conditional and regular LMs, the size of the training data
is important to achieve a low held-out perplexity. This is 
an obstacle for domains with limited available training data.
This issue has led to various model adaptation approaches.
These methods leverage large amounts of generic training data $D$ along with a 
small amount of target domain training data $T$ from the domain of interest.
{\it Fine tuning} is a popular domain adaptation method which trains a neural language
model in two phases, first maximizing the likelihood of the generic set $D$ 
(pre-training) before optimizing the likelihood of the target domain set $T$ 
(fine-tuning). 
As an alternative to fine-tuning, some methods consider leveraging the small 
target-domain training set to identify and emphasize similar data in the 
larger generic training set. These {\it emphasis} methods can be used 
individually or in conjunction with fine-tuning.
Emphasis methods include importance sampling, contrastive data selection and 
influence functions. This paper shows that these methods -- although 
proposed in different context -- can be presented in a unified way which 
allows light to be cast on their subtle differences. 

\section{Training Strategies}

This section first examines in-domain training, i.e. when the training and test data are sampled from the same distribution. It then studies out-of-domain training, i.e. when the training and test data distribution differs. Finally, it examines out-of-domain pre-training followed by in-domain fine tuning. For the three cases, we decompose the loss relying on classical concepts 
from learning theory and study the trade-offs involved in each setup.

\subsection{In-Domain Training}
\label{sec:in_dom}

Given a training set $D$ sampled from a distribution ${\cal D}$,
learning an LM typically aims at minimizing the 
negative log-likelihood of $D$, also referred to as the {\it cross-entropy}
loss i.e.
\begin{equation*}
\resizebox{1.0\hsize}{!}{$
\displaystyle
{\cal L}(\theta; D)
= - \frac{1}{|D|} \sum_{y \in D} \log P(y | \theta) 
=  \mathop{\mathbb{E}}_{y \sim D}[ -\log P(y | \theta)].
$}\end{equation*}
This {\it empirical risk} is the average over the finite set $D$, which acts as
a proxy for the expectation over the true, unavailable distribution $P(y| {\cal D})$,
\begin{eqnarray}
{\cal L}(\theta; {\cal D}) 
= - \sum_{y \in \Omega} \log P(y | \theta) P(y | {\cal D}) \nonumber\\
= \mathop{\mathbb{E}}_{y \sim {\cal D}}[ -\log P(y | \theta) ],\nonumber
\end{eqnarray}
where the distribution's support $\Omega$ is the set of all finite sequences.
The true expected loss is bounded by the entropy of the distribution $P(\cdot | {\cal D})$,
i.e.
$$
{\cal L}(\theta; {\cal D}) \ge {\cal L}_{H}({\cal D}) = H(P(\cdot | {\cal D})) 
$$
since $H(P(\cdot | {\cal D})) = \min_q \mathop{\mathbb{E}}_{y \sim {\cal D}}[ -\log q(y) ].$
The gap between the best likelihood from a neural network with the chosen 
parameterization and the entropy is called the approximation error 
$$
{\cal L}_{\rm app}({\cal D}, \Theta) 
= \min_{\theta \in \Theta} {\cal L}(\theta; {\cal D}) - H(P(\cdot | {\cal D})).
$$
This gap accounts for the fact that $P(\cdot | {\cal D})$ generally cannot be 
represented by a parameterized function from the chosen family spanned by $\Theta$.
In addition to the approximation error, one should consider the estimation
error to account that one relies on the empirical risk from the 
finite set $D$,
$$
{\cal L}_{\rm est}({\cal D}, \Theta, D) = {\cal L}(\theta_D; {\cal D})  - \min_{\theta} {\cal L}(\theta; {\cal D})
$$
with $\theta_D = \arg\min_{\theta\in\Theta} {\cal L}(\theta; D)$. Therefore, the
loss of $\theta_D$ over ${\cal D}$ decomposes as~\cite{DBLP:conf/nips/BottouB07}
\begin{empheq}[box=\widefbox]{multline}
{\cal L}(\theta_D; {\cal D})
= \\{\cal L}_{H}({\cal D}) + {\cal L}_{\rm app}({\cal D}, \Theta) + {\cal L}_{\rm est}({\cal D}, \Theta, D)
\label{eq:main_decomposition}
\end{empheq}
where the three terms accounts for 
the intrinsic uncertainty of ${\cal D}$, the chosen neural 
architecture and the finite training set $D$ respectively. 

The approximation error ${\cal L}_{\rm app}({\cal D}, \Theta)$ 
depends on the selected model family $\Theta$. It can be 
reduced by selecting a more expressive family, i.e. a
neural architecture with more capacity, a larger $\Theta$, e.g. 
architectures with more, wider layers. The estimation error
${\cal L}_{\rm est}({\cal D}, \Theta, D)$
depends both on the selected model family $\Theta$ and the size of the 
training data $D$. Increasing model capacity will result in a 
higher estimation error for the same training set size, but
training over a larger training set will decrease estimation error.
Therefore, for a given training set size, capacity needs to be chosen
to identify a good trade-off between the two error types. 

Two important properties of neural networks need to be kept in mind
when examining this trade-off. The {\it universal approximation} 
property~\cite{lecun87,funahashi1989approximate}
means that for any approximation error $\epsilon$ and any distribution 
${\cal D}$, there exists a capacity setting $C(\epsilon, {\cal D})$ at 
which a neural network $\theta \in C(\epsilon, {\cal D})$ whose error is below $\epsilon$, i.e.
$$
\forall \epsilon > 0, \exists~ C \textrm{~s.t.~}
{\cal L}_{\rm app}({\cal D}, C) \le \epsilon.
$$
In layman terms, the {\it universal approximation} property means
that for sufficiently large capacity settings, the approximation error 
can become arbitrary low.
The {\it statistical consistency} property means that for any 
$\epsilon, \epsilon' > 0$, there exist a training set size $N(\epsilon, {\cal D})$
such that sampling a training set of size $N(\epsilon, \epsilon', {\cal D})$ from
${\cal D}$ will result in an estimation error less than $\epsilon'$
with probability $1 - \epsilon$, $\forall \epsilon, \epsilon' > 0, \exists~ N {\rm~s.t~}$,
\begin{equation*}
P(D \sim {\cal D}^N : {\cal L}_{\rm est}({\cal D}, \Theta, D) < \epsilon' ) = 1 - \epsilon
\end{equation*}
In layman terms, the {\it statistical consistency} property means that 
for sufficiently large training sets, the probability to get an 
estimation error below any positive value can be arbitrary close to 1.

Universal approximation and consistency implies that, in the asymptotic case
(i.e. as the size of $D$ tends to infinity), the last two terms in 
Eq.~\ref{eq:main_decomposition} can be arbitrary close to zero with 
the appropriate model capacity (with high probability). In that case, 
the likelihood ${\cal L}(\theta_D; {\cal D})$ amounts to the intrinsic 
entropy of ${\cal D}$ with the appropriate model capacity.

\subsection{Out-of-Domain Training}
\label{sec:out_dom}

This section considers a setup where one needs a specialized language 
model in a domain ${\cal T}$ and two training sets are available:
a small training set T sampled from ${\cal T}$ and a large training set $D$
sampled from ${\cal D}$, a generic domain different from the specialized domain.

In that context, the simplest options are either to train a model over $T$
or $D$ alone. Training only on the small set $T$ results in the generalization 
loss
\begin{multline*}
{\cal L}(\theta_T; {\cal T})\\
= {\cal L}_{H}({\cal T}) + {\cal L}_{\rm app}({\cal T}, \Theta) + {\cal L}_{\rm est}({\cal T}, \Theta, T) 
\end{multline*}
with $\theta_T = \arg\min_{\theta\in\Theta} {\cal L}(\theta; T)$ as in the previous section.
Training on the larger set $D$ results in
\begin{multline*}
{\cal L}(\theta_D; {\cal T})\\
= {\cal L}_{H}({\cal T}) + {\cal L}_{\rm app}({\cal T}, \Theta) + {\cal L}_{\rm est}({\cal T}, \Theta, D).
\end{multline*}
Two factors are important to compare these two options: the 
size of the specialized set $T$ relative to the size of the 
generic set $D$ and the similarity between ${\cal T}$ and 
${\cal D}$ distributions.

When the ${\cal T}$ and ${\cal D}$ distributions are identical,
$D$ and $T$ are sampled from the same distribution and 
training a model on the larger training set $D$ is advantageous. 
For a constant capacity, this option will get a lower estimation 
error. When varying capacity, one might identify a setting with
an even better trade-off in the compound loss of 
Eq.~(\ref{eq:main_decomposition}) with the larger training set D.

When the distributions ${\cal T}$ and ${\cal D}$ differ and the 
size of D is fixed, the size of $T$ determines which option to prefer.
Statistical consistency means that 
${\cal L}_{\rm est}({\cal T}, \Theta, T)$
will converge to zero in probability as the size of $T$ grows.
This means that when the size of $T$ is greater than
$
N(\epsilon, {\cal L}_{\rm est}({\cal T}, \Theta, D), {\cal D}),
$
the probability that training on $T$ results in a better 
generalization loss than training on D is above \mbox{$1 - \epsilon$}.

When the distributions ${\cal T}$ and ${\cal D}$ differ, the 
Kullback–Leibler (KL) divergence between the two distributions
plays a key role.\\
\noindent {\bf Theorem 1}
The generalization of the 
loss of $\theta_D$ over {\cal T} is upper bounded as
\begin{empheq}[box=\widefbox]{multline}
\forall \epsilon > 0, ~ \exists N \textrm{~s.t.~} \forall D \sim {\cal D}^n,\\
{\cal L}(\theta_D; {\cal T}) \le H({\cal T}) + KL({\cal T}, {\cal D}) + \epsilon
\label{eq:out_of_domain_generalization_bound}
\end{empheq}
with probability $1 - \epsilon$. This bound justifies the
intuition that, if given the choice between two generic domains ${\cal D}$
and ${\cal D'}$, training over the one with the lowest KL divergence to ${\cal T}$
will result in a better asymptotic behaviour. The proof of this bound
is presented in Appendix~\ref{proof}.

\subsection{Fine-Tuning \& Multitask Learning}
\label{sec:fine_tuning}

Fine-tuning for domain adaptation trains a model on a small in-domain
set initializing optimization from the parameters of a model trained 
on a large out-of-domain set. 
Formally, fine-tuning minimizes ${\cal L}(\theta; T)$  the loss over T 
for a few steps, starting the optimization from 
\mbox{$\theta_D = \argmin_{\theta \in \Theta} {\cal L}(\theta; D).$} 
This strategy implicitly targets a trade-off between the empirical losses over $T$ and $D$. This trade-off is controlled by the number 
of fine tuning steps $n_{\rm ft}$. Few steps means that the identified 
parameters $\theta_{\rm ft}$ achieve a low loss over $D$, while many 
steps expresses that the parameters achieve a low loss over $T$. This strategy leverages the regularization effect of early stopping~\cite{caruana2001overfitting}, i.e. the solution found by gradient descent
is guaranteed to be in an Euclidean ball centered around the initialization whose
radius grows with the number of steps~\cite{grangier:2008:tpami}, i.e.
$$
\|\theta_{\rm ft} - \theta_D\|_2 \le \lambda ~ n_{\rm ft} ~g_{\max}
$$
where $\lambda$ refers to the (maximum) learning rate and $g_{\max}$
to an upper bound on the update norm.
The small distance between $\theta_{\rm ft}$ and $\theta_D$ guarantees
that the loss ${\cal L}(\theta_{\rm ft}; D)$ is close to the optimum
${\cal L}(\theta_D; D)$ when $\theta \to {\cal L}(\theta; D)$ is a 
smooth function, e.g. a Lipschitz function.

For the basic fine-tuning setup, several variants have been introduced. 
Some approaches~\cite{bert,liu2019roberta,t5} consider leaving some 
parameters un-tuned or frozen which is the extreme case of regularization for these weights, penalizing any deviation from initialization. Other approaches consider introducing novel (unregularized) weights for fine tuning, often referred 
as {\it adapter} 
layers~\cite{DBLP:journals/corr/abs-1902-00751,pmlr-v97-stickland19a,pfeiffer2020adapterfusion}.
Other forms of regularization, such as dropout, have also been considered in conjunction with fine tuning~\cite{miceli-barone-etal-2017-regularization}.

The selection of the regularization strength in fine-tuning is computationally efficient since it successively visits an optimization path from the most regularized 
model ($\theta_D$ trained only on D, Sec.~\ref{sec:out_dom}) to the unregularized $\theta_T$ (Sec.~\ref{sec:in_dom}). This is more efficient compared to explicit regularization methods, including multitask
learning~\cite{DBLP:books/sp/98/Caruana98,collobert08,pilault2021conditionally}, i.e. optimizing \mbox{$
{\cal L}_{\rm multi}(\theta; T, D, \alpha) 
= {\cal L}(\theta; T) + \alpha {\cal L}(\theta; D).
$}

\section{Data Selection}
\label{sec:data_selection}

Data selection aims to improve out-of-domain training by 
selecting or giving stronger weights to some data points.
The identification of these points aims to emphasize out-of-domain 
examples which have an impact on the model similar to the impact of the 
in-domain training examples. We study three independently 
proposed selection methods,
importance sampling, contrastive data selection and influence functions.
We show that these methods all train models through weighted log-likelihood training,
$$
{\cal L}(\theta; {D}, T, w) =\\ -\frac{1}{|D|} \sum_{y \in D} w(y; {\cal T, D}) \log P(y | \theta)
$$
but introduce their weights $w(y; {\cal T, D})$ with different justifications.
Despite these differences, we show that these methods result
in surprisingly similar selection weights in the specific case of neural language models.

Data selection is particularly suited when the out-of-domain training
distribution and the test distribution have a large KL divergence but the
out-of-domain training set is large. In that case, the generalization of a
model trained on out-of-domain data is poor due to the large KL divergence 
between ${\cal T}$ and ${\cal D}$, see
Eq.~(\ref{eq:out_of_domain_generalization_bound}).
When this KL divergence is large but out-of-domain 
data is abundant, data selection methods propose to select a
subset of the out-of-domain data $D^{\cal T} \subset D$. Ideally, 
the training loss over such a subset ${\cal L}(\theta, D^{\cal T})$ 
would be a better proxy for the generalization 
loss over ${\cal T}$, ${\cal L}(\theta, {\cal T})$, 
than the training loss over the full set $D$, ${\cal L}(\theta, D)$.

Selection involves a delicate trade-off though. One one hand, data selection is attractive since it replaces the training set with another set closer to the test domain. On the other hand, 
this training set is smaller, which increases the impact of 
estimation errors. Additionally, data selection is imperfect 
since the target domain distribution ${\cal T}$ is only 
known through a small target training set $T$. 

This section successively presents importance sampling, contrastive data 
selection and influence functions and connect them into a single framework.

\subsection{Importance Sampling}
\label{sec:importance}

Although intelligent selection also called contrastive data 
selection is more common~\cite{moore-lewis-2010-intelligent,wang-etal-2018-denoising}, we first examine importance sampling since this method will guide 
our understanding of other selection methods.

Importance sampling is a generic statistical technique~\cite{owen13}.
In our case, it can be used to estimate the expectation
of the cross-entropy loss over ${\cal T}$ while having
access to samples from ${\cal D}$. It relies on
the identity
\begin{align*}
{\cal L}(\theta; {\cal T}) 
& = && \E_{y \sim {\cal T}} [-\log P(y | \theta)]
\nonumber\\
& = && - \sum_{y \in \Omega} \log P(y | \theta) P(y | {\cal T})
\nonumber\\
& = && - \sum_{y \in \Omega} \log P(y | \theta) \frac{P(y | {\cal T})}{P(y | {\cal D})} P(y | {\cal D})
\nonumber\\
& = && \E_{y \sim {\cal D}} [-w(y; {\cal T, D}) \log P(y | \theta)]
\nonumber
\end{align*}
where $w(y; {\cal T, D}) = \frac{P(y | {\cal T})}{P(y | {\cal D})}$, 
assuming full support on ${\cal D}$, i.e. 
\mbox{$\forall y \in \Omega$}, $P(y | {\cal D}) > 0$.
In practice, one has not access to ${\cal T}$ and ${\cal D}$
but to finite samples $T$ and $D$. With importance sampling,
we can consider two alternative estimators of ${\cal L}(\theta; {\cal T})$, 
either the empirical risk over $T$,
$$
{\cal L}(\theta; {T}) = -\frac{1}{|T|} \sum_{y \in T} \log P(y | \theta)
$$
or the mean of the importance weighted cross entropy over $D$, i.e.
{\small
$$
{\cal L}_{\rm imp}(\theta; {D}, T, \hat{w}) =\\ -\frac{1}{|D|} \sum_{y \in D} \hat{w}(y; {\cal T, D}) \log P(y | \theta)
$$}%
where $\hat{w}$ estimates of the weights $w$ from the training sets $D$ and $T$.
The trade-off between these two estimators depends on 
the relative size of $T$ and $D$, the imbalance of the 
weights $w$ and the quality of their
estimate $\hat{w}$.

Importance sampling is interesting when the generalization
error ${\cal L}(\theta_{{\rm imp}(D, T)}; {\cal T})$ of the model 
$$\theta_{{\rm imp}(D, T)} = \argmin_\theta {\cal L}_{\rm imp}(\theta; D, T, \hat{w})$$
is less than the generalization error of $\theta_T$ selected by minimizing
${\cal L}(\theta; {T})$, i.e. classical empirical risk minimization.
This error decomposes as,
\begin{multline*}
{\cal L}(\theta_{{\rm imp}(D, T)}; {\cal T}) 
\\= 
{\cal L}_{H}({\cal T}) + {\cal L}_{\rm app}({\cal T}, \Theta) + {\cal L}^{\rm imp}_{\rm est}({\cal T}, \Theta, D, T).
\end{multline*}
We further decompose the estimation error in two terms,
\begin{multline*}
{\cal L}^{\rm imp}_{\rm est}({\cal T}, \Theta, D, T)\\ = 
{\cal L}_{\rm est/w}({\cal T}, {\cal D}, \Theta, D) 
+ {\cal L}_{\rm est/\hat{w}}({\cal T}, \Theta, D, T)
\end{multline*}
where ${\cal L}_{\rm est/w}({\cal T}, {\cal D}, \Theta, D)$ 
refers to the estimation error resulting from the finite size 
of $D$, assuming access to the true importance weights, 
and ${\cal L}_{\rm est/\hat{w}}({\cal T}, \Theta, D, T)$
isolate the residual error resulting from the estimation of $w$.
We have
\begin{multline*}
{\cal L}_{\rm est/w}({\cal T}, {\cal D}, \Theta, D) 
\\
= {\cal L}(\theta_{{\rm imp}(D, {\cal D})}; {\cal D}) - \min_{\theta} {\cal L}(\theta; {\cal T}),
\end{multline*}
\vspace{-10mm}
\begin{multline*}
{\cal L}_{\rm est/\hat{w}}({\cal T}, \Theta, D, T)
\\
= {\cal L}(\theta_{{\rm imp}(D, {\cal T})}; {\cal D}) - {\cal L}(\theta_{{\rm imp}(D, T)}; {\cal D})
\end{multline*}
with $\theta_{{\rm imp}(D, {\cal D})} = \argmin_\theta {\cal L}_{\rm imp}(\theta; D, T, \hat{w})$

The first term depends on the size of $D$ and the imbalance of 
the weights. For instance, if the weights are mostly concentrated 
over a small subset of $D$, this estimation error will be high. 
If this subset is smaller than
$T$, estimation errors from ${\cal L}_{\rm imp}(\theta; {D}, T, \hat{w})$
will be higher than from ${\cal L}(\theta; {T})$. The notion of 
{\it effective sample size} has been defined to quantify this effect~\cite{kish1965survey}.
It is defined by examining the variance of the weighted sum of
$n$ independent random variables $Z_i$ with mean $\mu_Z$ and variance $\sigma_Z^2$,
$
S_w = \frac{\sum_i w_i Z_i}{\sum_i w_i}.
$
This variance is
$$
\sigma^2_{S_w} = \frac{\sum_i w^2_i}{(\sum_i w)^2} \sigma^2_Z
$$
which can be compared to $\sigma^2_{S} = \frac{1}{n} \sigma^2_Z$
in the unweighted case. This means that the weighted sum variance
 matches the variance of an unweighted case with
$$
n_e =  \frac{(\sum_i w)^2}{\sum_i w^2_i}.
$$
Assuming that losses over ${\cal D}$ and ${\cal T}$ have comparable
means and variances, the expected loss estimate with importance 
weighting over $D$ has lower variance than the mean over $T$
only when,
$$
n_e = \frac{(\overline{w})^2}{\overline{w^2}} |D| \gg |T| 
$$
where $\overline{w} = \frac{1}{|D|} \sum_{y\in D} w(y)$
and $\overline{w^2} = \frac{1}{|D|} \sum_{y\in D} w^2(y)$
are the sample mean and variance of the weights over $D$.
This means that the first term in the estimation error is
${\cal L}_{\rm est/w}({\cal T}, \Theta, D, T)$ advantageous 
compared to the estimation error from classical empirical risk 
minimization over $T$ when $T$ is small.

Unfortunately, the second estimation error term 
${\cal L}_{\rm est/\hat{w}}({\cal T}, \Theta, D, T)$
gets larger as $T$ gets smaller since estimating the importance
weights 
$
w(y; {\cal T, D}) = \frac{P(y | {\cal T})}{P(y | {\cal D})}
$
from data is challenging when $T$ is small. One can remark that
language modeling is actually the very problem of identifying a model 
to estimate the probabilities in this ratio, 
$P(y | {\cal T})$ and $P(y | {\cal D})$, from finite samples 
from the distributions ${\cal T}$ and ${\cal D}$. Discriminative 
classifiers are also relevant to estimate this ratio since
$$
w(y; {\cal T, D}) \propto \frac{P({\cal T}|y)}{P({\cal D}|y)}.
$$
In fact the multiplying constant (prior ratio) does not matter
since multiplying the weighted loss by a positive constant has
no impact on optimization.

When importance weights are estimated with an LM,
one can estimate $P(\cdot |{\cal T})$ by fine tuning on $T$
a model pre-trained on $D$. The number of tuning steps $n_{\rm ft}$ 
gives controls on $\|\theta_{\rm ft} - \theta_D\|$. When $n_{\rm ft} = 0$, 
$\hat{w} = 1$ and 
the importance sampling loss corresponds to ${\cal L}(\theta, D)$. 
As $n_{\rm ft}$ grows, the estimate $P(y|\theta^{\rm ft})$ could 
overfit and assigns most of the probability mass to a small 
neighborhood around samples in $T$. The weights $\hat{w}$ will 
in turn be concentrated in this small neighborhood, making the 
minimizer of the importance sampling loss close to the minimizer 
of the empirical loss over $T$. Therefore, fine-tuning a language
model for estimating the importance weights allow to progressively 
transition between the in-domain and the out-of-domain empirical 
loss minimizers seen in Section~\ref{sec:out_dom}. In the next sections,
we refer to the estimated importance sampling weights as
$$
w^{\rm imp}_{D, T}(y) = \hat{w}(y; T, D).
$$
Importance sampling has been used for model training for various application: 
either to improve training speed~\cite{johnson2018training,katharopoulos2018}
or to adapt to a changing training distribution~\cite{mahmood2014weighted,metelli2018policy}.
Importance sampling has rarely been used to modify the training distribution of language 
models~\cite{foster-etal-2010-discriminative,fernandez-downey-2018-sampling} as intelligent selection methods are more common.

\subsection{Intelligent Selection}
\label{sec:cds}

Intelligent selection~\cite{moore-lewis-2010-intelligent,axelrod-etal-2011-domain} and contrastive data 
selection, its extension to neural networks~\cite{van-der-wees-etal-2017-dynamic,
wang-etal-2018-denoising}, have 
been introduced in the language modeling literature. We show that these methods are 
closely related to importance sampling, even if their original papers does not mention 
this link. 

Intelligent selection selects training samples from an out-of-domain dataset 
according to the log-odd between an in-domain LM and an out-of-domain 
LM. Typically, a binary decision is taken per sentence by comparing 
the average log-odd to a threshold $\tau$,
$$
{\cal L}^{\rm IntSel}(\theta, D, T) 
= -\sum_{y \in D} b^{\rm IntSel}_{D, T}(y)
\log P(y | \theta)
$$
where $
b^{\rm IntSel}_{D, T}(y)$ is defined as 
${\rm I}\left\{\log P(y | \theta_T) - \log P(y | \theta_D) > \tau\right\}
$.
Compared to importance sampling, the weights are binarized, i.e.
\begin{empheq}[box=\widefbox]{equation*}
b^{\rm IntSel}_{D, T}(y) = 
{\rm I}\left\{\log w^{\rm imp}_{D, T}(y) > \tau \right\}.
\end{empheq}
The binarization decision was 
certainly driven by convenience as most n-gram LM training 
packages did not support weighted likelihood optimization when intelligent
selection was introduced. Binarization 
also has the advantage of down-weighting extremely positive weight
values from large $\log P(y | \theta_T)$ due to over-fitting on the small 
set $T$. 

More recently, intelligent selection has been extended to neural 
models~\cite{van-der-wees-etal-2017-dynamic,wang-etal-2018-denoising}. 
Contrastive data selection~\cite{wang-etal-2018-denoising} 
suggests to fine tune the in-domain model $\log P(y | \theta_T)$ 
from $\log P(y | \theta_D)$ and also observes that 
selection scores can efficiently be estimated from a model with a much smaller 
capacity than the final trained model.
Dynamic selection~\cite{van-der-wees-etal-2017-dynamic} proposes 
to increase the selection threshold $\tau_t$ as training progresses, gradually 
transitioning from generic to in-domain training. This gradual 
adaptation of neural network is related to curriculum 
learning~\cite{DBLP:conf/icml/BengioLCW09} which studies 
the ordering of examples and tasks during model training. 

Intelligent selection methods have been applied both for unconditional models 
(language modeling) and conditional models (machine translation).
In the conditional case, intelligent selection computes
\begin{multline*}
b^{\rm IntSel}_{D, T}(x, y) = {\rm I}\left\{\log w^{\rm IntSel}_{D, T}(x, y)  > \tau\right\}
\\\textrm{with}\quad
w^{\rm IntSel}_{D, T}(x, y) = \frac{P(y | x, \theta_T)}{P(y | x, \theta_D)}.
\end{multline*}
This ratio of conditional probabilities is different from the ratio of joint probabilities 
stemming from importance sampling, i.e.
\begin{multline*}
{\cal L}_{\rm imp}(\theta; {D}, T, \hat{w}) 
= \\
-\frac{1}{|D|} \sum_{y \in D} \frac{P(x, y | {\cal T})}{P(x, y | {\cal D})} \log P(y | x, \theta).
\end{multline*}
The two ratios match when $P(x|{\cal T})=P(x|{\cal D})$ since
\begin{empheq}[box=\widefbox]{align*}
w^{\rm imp}_{D, T}(x, y) 
&=&& \frac{P(x, y | {\cal T})}{P(x, y | {\cal D})}\\ 
&=&& \frac{P(x|{\cal T})}{P(x|{\cal D})} ~ w^{\rm IntSel}_{D, T}(x, y).
\end{empheq}
The formulation of intelligent selection therefore neglects the 
domain mismatch from the input distribution in the conditional case.
This formulation aligns with the denoising goal~\cite{wang-etal-2018-denoising}
which assumes that $D$ contains label noise, i.e. mistranslation in that case.

\subsection{Influence Functions}
\label{sec:influence}

As mentioned above, importance sampling and intelligent selection weights
can be estimated by contrasting the log probabilities from a base model 
with those from a fine-tuned model. This use of fine-tuning connects
intelligent selection to influence function and gradient alignment 
techniques. 
Influence functions~\cite{koh-liang-influence-17,DBLP:conf/nips/PruthiLKS20} 
have been used as a diagnostic tool to identify the training instances which support or 
contradict with a given test label. This task is related to the selection of training 
data when the objective is to find instances in a generic training set
$D$ whose training updates increase the likelihood of a set $T$ from a different domain.

The influence of a training point $y$ on a test point $y'$ 
is defined as 
$$
{\rm I}(y, y') = 
-  \frac{\partial \ell}{\partial \theta}(y'; \theta)^\top
H^{-1}_\theta
 \frac{\partial \ell}{\partial \theta}(y; \theta)  
$$
where $\ell(y, \theta)$ refers to the loss at $y$ for a model with parameters $\theta$
and $H_\theta$ refers to the Hessian of the model loss at $\theta$. This quantity can be
derived by considering the impact of reducing the weight of point $y$ during training
on the test loss at $y'$.
If we increase the weight of a training example by $\epsilon$,
$$
\theta_{D, \epsilon} = \min_\theta \frac{1}{|D|}\sum_{z \in D} \ell(z; \theta) + \epsilon  \ell(y; \theta)
$$
From~\cite{cook1982residuals}, we derive
$$
\left. \frac{\partial \theta_{D, \epsilon} }{\partial \epsilon}\right|_{\epsilon=0} =   - H^{-1}_\theta
\left. \frac{\partial \ell}{\partial \theta}(y; \theta)\right|_{\theta=\theta_D}.  
$$
Composing with the test loss on $(x', y')$, we get
\begin{small}$$
\left. \frac{\partial \ell(y'; \theta_{D, \epsilon})}{\partial \epsilon}  \right|_{\epsilon=0} = - 
\left. \frac{\partial \ell(y'; \theta)^\top}{\partial \theta}\right|_{\theta=\theta_D}
H^{-1}_\theta
\left. \frac{\partial \ell(y; \theta)}{\partial \theta}\right|_{\theta=\theta_D}
$$\end{small}%
which matches the expression of influence introduced above.

We now connect influence with the precedent sections on importance sampling and 
contrastive data selection. We consider an LM with weights $\theta_D$, 
trained on the generic 
training set $D$. Its first order Taylor expansion at $\theta_D$ is
\begin{multline}
\log P(y|\theta_D + \Delta \theta) =\\
\log P(y|\theta_D)
+ \Delta \theta^\top g(y; \theta_D)
+ O\left( \|\Delta \theta\|^2 \right)
\label{eq:taylor}
\end{multline}
where $g(y; \theta_D) = \left. \frac{\partial}{\partial \theta}\log P(y | \theta)  \right|_{\theta=\theta_D}$.
If the model pre-trained on $D$ is fine-tuned on $T$ by performing a
single step of gradient descent with learning rate $\lambda$, we get
\begin{eqnarray*}
\theta_T 
& = & 
\theta_D - \lambda \left. \frac{\partial}{\partial \theta} {\cal L}(T; \theta) \right|_{\theta=\theta_D} \\
& = &
\theta_D + \lambda \E_{y \sim T}\left[ g(y; \theta_D) \right].
\end{eqnarray*}
In that case, the log-odd of the two models therefore has the following Taylor expansion,
\begin{multline*}
\log P(y|\theta_T) - \log P(y|\theta_D)\\
= \lambda \E_{y' \sim T}\left[  g(y'; \theta_D)^\top g(y; \theta_D) \right]\\
+ O\left( \|\theta_D - \theta_T\|^2 \right).
\end{multline*}
If we assume that the model's Hessian is the identity, $H_\theta = \mathbbm{1}$, we therefore have
\begin{multline*}
\log P(y|\theta_T) - \log P(y|\theta_D)
= \\
-\lambda \E_{y' \sim T}\left[ {\rm I}(y, y') \right]
+ O\left( \|\theta_D - \theta_T\|^2 \right).
\end{multline*}
The Hessian assumption might be dropped when the model is fine-tuned
with a Newton-style update~\cite{DBLP:books/cu/BV2014}. 
The above relation means that the negative mean influence of a point $y \in D$ over the set $T$
also corresponds to the log of the estimated importance weights introduced in 
Section~\ref{sec:importance}, i.e.
\begin{empheq}[box=\widefbox]{multline*}
\log w^{\rm imp}_{D, T}(y)
= \\
-\lambda \E_{y' \sim T}\left[ {\rm I}(y, y') \right]
+ O\left( \|\theta_D - \theta_T\|^2 \right).
\end{empheq}
Of course, this relation holds only in the case where a single gradient step is
performed for fine-tuning.
This relation allows estimating the reduction in test loss 
(here over $T$) when removing training samples from $D$ with positive 
influence which is also the goal of intelligent data selection.
This strategy has been applied to label noise 
filtering~\cite{koh-liang-influence-17}, class rebalancing~\cite{ren2018learning} and domain 
adaptation~\cite{wang2021gradientguided}.

\subsection{Comparing Data Selection Methods}

Our analysis connects importance sampling, contrastive data selection and influence functions. In practice, contrastive data selection is the most popular approach. Unlike influence functions, contrastive data selection weights rely on fine tuning the generic model for more than one step on the in-domain data $T$. This has two effects. On one hand the contrastive data selection weights can be more reliable, closer to the ideal weights $w(y; {\cal T, D}) = \frac{P(y | {\cal T})}{P(y | {\cal D})}$. On the other hand, multiple steps increase the risk of over-fitting to $T$. In the case where one first trains with data selection before fine tuning on $T$, it might actually be helpful to limit the influence of $T$ on selected data, to increase the complementary effect of fine tuning~\cite{iter:complementarity}.

When comparing contrastive data selection with importance sampling, the weight binarization is the main difference. This binarization might also have two opposite effects. On the positive side, it acts has a regularizer since binary weights are less likely to reflect statistics specific to $T$ compared to unquantized ones. On the negative side, it cancels low weights which might collectively represent most of the weighted cross entropy. This interpretation of contrastive data selection as a regularized version of importance sampling opens the door to exploring more sophisticated regularization alternative to regularization, e.g. using a lower capacity model or different input features to estimate selection weights.

\section{Conclusions}

This work focuses on domain adaptation for neural language modeling. It compares the generalization 
properties of a model trained over a large out-of-domain corpus as opposed to a model trained
over a small in-domain corpus. It shows how fine-tuning, the most common approach for neural LM
adaptation can achieve better trade-offs than either solution. We then focus on adaptation via data selection techniques, i.e. techniques to emphasize in-domain data in an out-of-domain training set. We show that common techniques, contrastive data selection and influence function selection, can both be derived from importance sampling.

Our analysis currently assumes a pure language modeling setup, i.e. an auto-regressive model trained aiming high log-likelihood, both for out-of-domain and in-domain data. In the future, we want to extend our analysis of domain adaptation techniques to the popular setting~\cite{foundation21} where model training combines language modeling over out-of-domain data and a different final task on in-domain data. 

Our theoretical work also raises empirical questions. The binarization of importance sampling weights in intelligent selection is a simple variance reduction technique and more sophisticated alternative might be beneficial empirically. The link between influence functions and importance sampling suggests that examples with importance sampling weights lower than one have only a negative effect on the in-domain likelihood, which is not a typical observation in practice. This contradiction suggests expanding influence scores to take into account effects beyond a single update.

\section*{Acknowledgements}

We thanks Wei Wang, Bowen Liang, Kelvin Guu and Nicolas Le Roux for their suggestions and comments.

\bibliographystyle{acl_natbib}
\bibliography{main}

\newpage
\appendix
\section{Proof of Theorem 1\label{proof}}

When the distributions ${\cal T}$ and ${\cal D}$ differ, the 
Kullback–Leibler (KL) divergence between the two distributions 
is considered. We show that the generalization of the 
loss of $\theta_D$ over {\cal T} is upper bounded
\begin{multline}
\forall \epsilon > 0, ~ \exists N \textrm{~s.t.~} \forall D \sim {\cal D}^n,\\
{\cal L}(\theta_D; {\cal T}) \le H({\cal T}) + KL({\cal T}, {\cal D}) + \epsilon
\end{multline}
with probability $1 - \epsilon$
This bound justifies
intuition that if given the choice between two generic domain ${\cal D}$
and ${\cal D'}$, training over the one with the lowest KL divergence to ${\cal T}$
will result a in better asymptotic behaviour.

\begin{proof} We consider the 
asymptotic case for the size of $D$. For any $\epsilon > 0$, 
the universal approximation property allows us to consider a 
model capacity large enough such that
$$
{\cal L}_{\rm app}({\cal D}, \Theta) < \frac{\epsilon}{2}
$$
Using consistency, we can also consider a training set $D$
large enough such that
$$
{\cal L}_{\rm est}({\cal D}, \Theta, D) < \frac{\epsilon}{2}
$$
with probability $1-\epsilon$. With the same probability,
$$
{\cal L}(\theta_D; {\cal D}) < {\cal L}_{H}({\cal D}) + \epsilon
$$
which can be rewritten as a bound on the Kullback-Leibler divergence,
$$
KL(P(\cdot | {\cal D}), P(\cdot | \theta_D)) = 
{\cal L}(\theta_D; {\cal D}) - {\cal L}_{H}({\cal D}) < \epsilon.
$$
This bound can help connecting the generalization loss of $\theta_D$
over {\cal T} with the Kullback-Leibler divergence of {\cal T} and {\cal D},
{\begin{small}
\begin{eqnarray}
& & {\cal L}(\theta_D; {\cal T})  \nonumber\\
& = & \sum_{y \in \Omega} P(y | {\cal T}) \log P(y | \theta_D)
\nonumber\\
& = & \sum_{y \in \Omega} P(y | {\cal T}) \log( P(y | {\cal D}) + P(y | \theta_D) - P(y | {\cal D}))
\nonumber\\
& \le & \sum_{y \in \Omega} P(y | {\cal T}) \log( P(y | {\cal D}) + | P(y | {\cal D}) - P(y | \theta_D)|)
\nonumber\\
& \le & \sum_{y \in \Omega} P(y | {\cal T}) \log( P(y | {\cal D}) + 2 \epsilon^2)\label{pinkser}\\
& \le & \sum_{y \in \Omega} P(y | {\cal T}) \log( P(y | {\cal D})) + \log( 1 + 2 m \epsilon^2)\nonumber\\
& \le & H({\cal T}) + KL({\cal T}, {\cal D}) + \log( 1 + 2 m \epsilon^2)\nonumber
\end{eqnarray}
\end{small}}%
where $m = 1 / \min_y  P(y | {\cal D})$ assumes that $P(\cdot |{\cal D})$
has full support, and (\ref{pinkser}) relies on Pinsker's inequality, i.e. 
$\max_y |P(y) - Q(y)| < 2 KL(Q, Y)^2$.
\end{proof}

\end{document}